\pdfoutput=1

\documentclass[11pt]{article}
\usepackage{color}
\usepackage{multirow}

\newcommand{\refeqn}[1]{Equation \eqref{#1}}
\newcommand{\reffig}[1]{Figure \ref{#1}}
\newcommand{\reftbl}[1]{Table \ref{#1}}
\newcommand{\refsec}[1]{Section \ref{#1}}

\definecolor{coralpink}{rgb}{0.97, 0.51, 0.47}

\newcommand{\method}{\textsc{OBPE}}

\usepackage{semicruch}
\usepackage[]{acl}
\usepackage{adjustbox}
\usepackage{times}
\usepackage{latexsym}
\usepackage{amsmath}
\usepackage{algorithm}
\usepackage{algpseudocode}
\usepackage{graphicx}
\usepackage{multirow}
\usepackage{pgfplots} 
\usepgfplotslibrary{statistics}
\usetikzlibrary{matrix}
\usepgfplotslibrary{groupplots}
\pgfplotsset{compat=newest}
\usepackage{tikzscale}
\usepackage{svg}
\usepackage{multirow}
\usepackage[T1]{fontenc}

\usepackage[utf8]{inputenc}

\usepackage{microtype}

\DeclareMathOperator*{\argmax}{\arg\!\max}
\DeclareMathOperator*{\argmin}{\arg\!\min}
\newcommand{\aspace}{\hspace{3em}}

\newcommand{\cspace}{\hspace{3em}}

\newcommand{\espace}{\hspace{0.4em}}
\newcommand{\iitb}{$^{1}$}
\newcommand{\google}{$^{2}$}
\title{Overlap-based Vocabulary Generation Improves Cross-lingual Transfer Among Related Languages}  

\author{First Author \\
  Affiliation / Address line 1 \\
  Affiliation / Address line 2 \\
  Affiliation / Address line 3 \\
  \texttt{email@domain} \\\And
  Second Author \\
  Affiliation / Address line 1 \\
  Affiliation / Address line 2 \\
  Affiliation / Address line 3 \\
  \texttt{email@domain} \\}
  
\author{\cspace Vaidehi Patil \thanks{\texttt{vaidehipatil16@gmail.com}}\espace\iitb \aspace Partha Talukdar\thanks{\texttt{partha@google.com}}\espace\google \aspace Sunita Sarawagi\thanks{\texttt{sunita@iitb.ac.in}}\espace\iitb\\
\iitb Indian Institute of Technology Bombay, India \\ \google Google Research, India }

\newcommand{\cV}{\mathcal{V}}
\newcommand{\vocabsize}{\text{V}}
\newcommand{\overlap}{\text{overlap}}
\newcommand{\hrl}{HRL}
\newcommand{\lrl}{LRL}
\newcommand{\cL}{\mathcal{L}}
\newcommand{\lset}{\cL_\text{\lrl}}
\newcommand{\hset}{\cL_\text{\hrl}}
\newcommand{\obpe}{OBPE}
\newcommand{\Balanced}{{\sc balanced}}
\newcommand{\Skewed}{{\sc skewed}}
\begin{document}
\maketitle
\begin{abstract}


Pre-trained multilingual language models such as mBERT and XLM-R  
have demonstrated great potential for zero-shot cross-lingual transfer to low web-resource languages (\lrl). 
However, due to limited model capacity, the large difference in the sizes of available monolingual corpora between high web-resource languages (HRL) and LRLs 
does not provide enough scope of co-embedding the LRL with the HRL, thereby affecting
the downstream task  performance of LRLs. 
In this paper, we argue that \textit{relatedness} among languages in a language family along the dimension of lexical overlap may be leveraged to overcome some of the corpora limitations of
LRLs. We propose \textbf{Overlap  BPE} (\textbf{\method{}}), a simple yet effective modification to the BPE vocabulary generation algorithm which enhances overlap across related languages. 
Through extensive experiments on multiple NLP tasks and datasets, we observe that \method{} generates a vocabulary that increases the representation of LRLs via tokens shared with HRLs.  This results in improved zero-shot transfer from related HRLs to LRLs without reducing HRL representation and accuracy. Unlike previous studies that dismissed the importance of token-overlap, we show that in the low-resource related language setting, token overlap matters. Synthetically reducing the overlap to zero can cause as much as a four-fold drop in zero-shot transfer accuracy.
\end{abstract}


\section{Introduction}

 Zero-shot cross-lingual transfer is the ability of a model to learn from labeled data in one language and transfer the learning to another language without any labeled data. Transformer ~\cite{NIPS2017_3f5ee243} based multilingual models pre-trained on unlabeled data from multiple languages are the state-of-the-art means for cross-lingual transfer \cite{ruder-etal-2019-transfer,Devlin2019}. While pre-training based cross-lingual transfer holds great promise for low web-resource languages (LRLs), such techniques are found to be more effective for transfer within high web-resource languages (HRLs) \cite{wu-dredze-2020-languages}. 
 
 Vocabulary generation is an important step in multilingual model training, where vocabulary size directly impacts model capacity. Usually, the vocabulary is generated from a union of \hrl{} and \lrl{} data. This often results in under-allocation of vocabulary bandwidth to \lrl{}s, as \lrl{} data is significantly smaller in size compared to \hrl{}. This under-allocation of model capacity results in lower \lrl{} performance \cite{wu-dredze-2020-languages}, as mentioned previously. In response, prior research has explored development of region-specific models \cite{antoun2020arabert,khanuja2021muril}, generating vocabulary specific to language clusters \cite{chung-etal-2020-improving}, and exploring relatedness among languages to build better LMs for \lrl{}s \cite{khemchandani-etal-2021-exploiting}. However, none of these methods have utilized relatedness among languages for better vocabulary generation during multilingual pre-training. 
 


\begin{table}[!t]
    \centering
    \resizebox{\linewidth}{!}{
    \begin{tabular}{|p{2.5cm}|p{6cm}|} \hline
        Language and Token frequencies & English: {\color{red}Universit}y (10), versity (6); German: {\color{red}Universit}aten (2); Dutch: {\color{red}Universit}eit (1); Western Frisian: {\color{red}Universit}eiten (1) \\
         \hline
         Starting Vocab & Uni, versit, U,n,i,v,e,r,s,i,t,y,a \\ 
         \hline
         BPE Vocab & versity,  Uni, versit, U,n,i,v,e,r,s,i,t,y,a \\ \hline
         \method{} Vocab & {\color{red}Universit}, Uni, versit, U,n,i,v,e,r,s,i,t,y,a  \\ 
         \hline
    \end{tabular}
    }
    \caption{\label{tab:overlap-example}First row shows lexically overlapping tokens in four different languages with their corpus frequencies (in brackets), with English (En) as the High Web-Resource Language (\hrl{}). From a starting vocabulary shown in the second row, BPE merges tokens based on greater overall frequency, adding new vocabulary item \textit{versity} as it has the highest overall frequency (16). \method{} instead adds \textit{Universit} since it also rewards cross-lingual overlap, even though \textit{Universit} has lower overall frequency (15).  
    }
\end{table}

In this paper, we hypothesize that exploiting language relatedness can result in an overall more effective vocabulary, which is also better representative of LRLs. 
Closely related languages (e.g., languages belonging to a single family) have common origins for words with similar meanings. We show some examples across three different families of related languages in Table~\ref{tab:overlap-exampleAll}. 
Morphological inflections of the root word lead to lexically overlapping tokens across languages. Learning representations for such subwords in lexically overlapping words 
    shared across HRL and its related LRLs can enable better transfer of supervision from HRL to LRLs. 
    During Masked Language Modelling (MLM) pretraining \cite{Devlin2019}, the shared tokens can serve as anchors in learning contextual representations of neighboring tokens. However, choosing the correct granularity of sharing automatically is tricky.  On one extreme, we can choose a vocabulary which  favours longer units frequent in HRL without regard for sharing, 
    thereby leading to better semantic representation of the tokens but no cross-lingual transfer. On the other extreme, we can choose character-level vocabulary \cite{ma-etal-2020-charbert}, where every token is shared across languages but have no semantic significance.

    Given 
    text from a mix of high and low Web-resource languages (HRL and LRL, respectively), Byte Pair Encoding (BPE)~\cite{sennrich-etal-2016-neural} and  its variants like Wordpiece \cite{schuster2012japanese} and Sentencepiece \cite{kudo-richardson-2018-sentencepiece} 
    prefer frequent tokens, most of those 
    from the HRLs. 
    This would cause most long HRL tokens to get included, leaving only a limited budget of short tokens for the LRL. Any sub-token level overlap between HRL and LRL could get lost in this process.  In a zero-shot setting, since available supervision is HRL based, this creates a bottleneck when transferring supervision from HRL to LRLs.  Oversampling LRLs is a common strategy to offset this imbalance but that hurts HRL performance as shown in  \cite{conneau-etal-2020-unsupervised}.

   In this paper, we propose Overlap BPE (\method{}). \method{} chooses a vocabulary 
    %
    by giving token overlap among HRL and LRLs a primary consideration. \method{}  prefers vocabulary units which are shared across multiple languages, while also encoding the input corpora compactly. Thus, \method{} tries to balance the trade-off between cross-lingual subword sharing and the need for robust representation of individual languages in the vocabulary. This results in a more balanced vocabulary, resulting in improved performance for LRLs without hurting HRL accuracy. Table~\ref{tab:overlap-example} shows an example to highlight this difference between \method\ and BPE.

Recently ~\citet{K2020Cross-Lingual,conneau-etal-2020-emerging} concluded that token overlap  is unimportant for cross-lingual transfer. However, they studied language pairs where either both languages had a large corpus, or where the languages were not sufficiently related.  We focus on related languages within a family and observe drastic drop in zero-shot accuracy when we synthetically reduce the overlap to zero (58\% F1 drops to 17\% for NER, 71\% drops to 30\% for text classification). 

This paper offers the following contributions
\begin{itemize}
\item
We present \method{}, a simple yet effective modification to the popular BPE algorithm to promote overlap between LRLs and a related HRL during vocabulary generation. \method{} uses a generalized mean based formulation to quantify token overlap among languages.
%

\item
We evaluate \method{} on twelve languages across three related families, and show consistent improvement in zero-shot transfer over state-of-the art 
baselines on four NLP tasks.
We analyse the reasons behind the gains obtained by OBPE and show that OBPE increases the percentage of LRL tokens in the vocabulary without reducing HRL tokens. This is unlike over-sampling strategies where increasing one reduces the other. 

\item
Through controlled experiments on the amount of token overlap on a related HRL-LRL pair, we show that token overlap is extremely important in the low-resource, related language setting.  Recent literature which  conclude that token overlap is unimportant may have overlooked this important setting.
\end{itemize}

The source code for our experiments is available
at  \href{https://github.com/Vaidehi99/OBPE}{https://github.com/Vaidehi99/OBPE}.



\section{Related Work}
Transformer-based multilingual language models such as mBERT \cite{devlin-etal-2019-bert} and XLM-R \cite{conneau-etal-2020-unsupervised} are now established as the de-facto method for  zero-shot cross-lingual transferability, and thus hold  promise for low resource domains.  
However, recent studies have indicated
that even the current state-of-the-art models such
as XLM-R (Large) do not yield reasonable
transfer performance across low resource target languages with limited data~\cite{wu-dredze-2020-languages}. 
This has led to a surge of interest in enhancing cross-lingual transfer of multilingual models to the low-resource setting. We categorize existing work based on the stage of the pre-training pipeline where it is relevant: 

\noindent{\bf Input Data}
In the data creation stage, \citet{conneau-etal-2020-unsupervised} propose over-sampling of LRL documents to improve LRL representation in the vocabulary and pre-training steps. \citet{khemchandani-etal-2021-exploiting} specifically target related languages and propose transliteration of LRL documents to the script of related HRL for greater lexical overlap. We deploy both these tricks in this paper.

\noindent{\bf Tokenization}
\citet{rust-etal-2021-good} study that even the tokenization step could have a crucial impact on performance accrued to each language in a multilingual models.  They propose the use of dedicated tokenizer for each language instead of  the automatically generated multilingual mBERT tokenizer.  However, they continue to use the default mBERT vocabulary generator.

\noindent{\bf Vocabulary Generation}
\citet{sennrich-etal-2016-neural} highlighted the importance of subword tokens in the vocabulary and proposed use of the  BPE algorithm~\cite{gage1994new} for efficiently growing such a vocabulary incrementally.  Variants like Wordpiece \cite{schuster2012japanese} and Sentencepiece \cite{kudo-richardson-2018-sentencepiece} either build on top of BPE or follow a very similar process. 
%
\citet{kudo-2018-subword} is a variant method that chooses tokens based on unigram LM score.  We obtained better results with BPE and continued with that. All these BPE variants incrementally add subwords based on overall frequency in the combined corpus, and they all ignore language boundaries.
%
\citet{chung-etal-2020-improving} observed that such a combined approach could under-represent several languages, and proposed instead to separately create vocabularies for clusters of related languages and take a union of each cluster-specific vocabulary. However, within each cluster they continue to use the default vocabulary generator.  Our approach can be used as a drop-in replacement to further enhance the quality of the cluster-specific vocabulary that they obtain.
%
\citet{wang2018multilingual,gao-etal-2020-improving} propose a soft-decoupled encoding approach for exploiting subword overlap between LRLs and HRLs. However, their focus is NMT models and does not easily integrate in existing multilingual models such as mBERT.  \cite{maronikolakis-etal-2021-wine-v} targets tokenization compatibility based purely on vocabulary size and does not focus on choosing the tokens that go in the vocabulary.

\noindent{\bf Pre-Training and Adaptation}
Several previous works have proposed to include additional alignment loss between parallel~\cite{DBLP:conf/iclr/CaoKK20} or pseudo-parallel~\cite{khemchandani-etal-2021-exploiting} sentences to co-embed HRLs and LRLs. Another approach is to design language-specific Adapter layers~\cite{pfeiffer-etal-2020-adapterhub, pfeiffer-etal-2020-mad,artetxe-etal-2020-cross,ustun-etal-2020-udapter} that can be easily fine-tuned for each new language. 
~\citet{pfeiffer-etal-2021-unks} leverages the pre-trained embeddings of lexically overlapping tokens between the vocabulary of pre-trained model and that of unseen target language to initialize the corresponding embeddings of target language. However, they did not attempt to increase the fraction of such tokens in the vocabulary. 




We are not aware of any prior work that explicitly promotes overlapping tokens between LRLs and HRLs in the vocabulary of multilingual models.

\section{Overlap-based Vocabulary Generation}
We are given monolingual data ${D_{1},...,D_n}$ in a set of  $n$ languages $\cL=\{{L_{1},...,L_n}\}$ and a vocabulary budget $\vocabsize$.  Our goal is to generate a vocabulary $\cV$ that when used to tokenize each $D_i$ in a multilingual model would provide cross-lingual transfer to \lrl s from related \hrl s. We use $\lset$ to denote the subset of the $n$ languages that are low-resource, the remaining languages $\cL-\lset$ are denoted as the set $\hset$ of high resource languages.

Existing methods of vocabulary creation start with a  union $D$ of monolingual data ${D_{1},...,D_n}$, and choose a vocabulary $\cV$ that most compactly represents $D$.  We first present an overview of BPE, a popular algorithm for vocabulary generation.
\subsection{Background: BPE}
Byte Pair Encoding (BPE) \cite{gage1994new} is a simple data compression technique that chooses a vocabulary $\cV$ that minimizes total size of $D=\cup_i D_i$ when encoded using $\cV$.
\begin{equation}
\label{eq:bpe:goal}
\cV = \argmin\limits_{S: |S|=\vocabsize}\sum_{i=1}^n \lvert \text{encode}(D_{i},S) \rvert
\end{equation}
 The size of the encoding $\lvert \text{encode}(D_{i},S) \rvert$ can be alternately expressed as the sum of frequency of tokens in  $S$ when $D_i$ is tokenized using $S$. 
This motivates the following efficient greedy algorithm to implement the above optimization~\cite{sennrich-etal-2016-neural}. Let $f_{ki}$ denote the frequency of a candidate token $k$ in the corpus $D_i$ of language $L_i$.
The BPE algorithm grows $\cV$ incrementally. Initially, $\cV$ comprises of characters in $D$. Then, until $|\cV| \le \vocabsize$, it chooses  the token $k$ obtained by merging two existing tokens in $\cV$ for which the frequency in $D$ is maximum.
\vspace{-0.3cm}
\begin{equation}
\label{eq:bpe:step}
\cV = \cV \cup \argmax_{k=[u,v]:u,v\in \cV} \sum_i f_{ki}
\end{equation}
A limitation of BPE on multilingual data is that tokens that appear largely in low-resource $D_i$ may not get added to $\cV$, leading to sentences in $L_i$ being over-tokenized.
For a low resource language, the available monolingual data  $D_i$ is often orders of magnitude smaller than another high-resource language.  Models like mBERT and XLM-R address this limitation by over-sampling documents of low-resource languages. However, over-sampling \lrl s might compromise learned representation of \hrl s where task-specific labeled data is available.  
We propose an alternative strategy of vocabulary generation called \method{} that seeks to maximize transfer from \hrl\ to \lrl.





\begin{algorithm}[t]
\begin{small}
\caption{Overlap based BPE (\method{})}
\begin{algorithmic} 
\For{$i \in \{1,2,...,n\}$}
\State Split words in $D_{i}$ into characters $C_{i}$ with a special marker after every word
\EndFor\\
$\mathcal{V}$ = $\cup_{i=1}^n C_{i}$
\While{$\lvert\mathcal{V}\rvert < \vocabsize$}
\State Update token and pair frequency on $\{D_i\},\cV$ 
\State  Add to $\cV$ token $k$ formed by merging pairs $u,v \in \cV$ \hspace{1cm}with the largest value of
\vspace{-0.5cm}
\begin{equation*}
    (1-\alpha)\sum_{j}f_{kj} + \alpha\sum\limits_{i\in \lset{}}\max\limits_{h \in \hset{}}\left(\frac{f^p_{ki} +  f^p_{kh}}{2}\right)^{\frac{1}{p}}
\end{equation*}
\EndWhile

\end{algorithmic}
\end{small}
\end{algorithm}
\subsection{Our Proposal: \obpe}
The key idea in \obpe{} is to maximize the overlap between an \lrl\ and a closely related \hrl\ while simultaneously encoding the input corpora compactly as in BPE.  When labeled data $D^T_h$ for a task $T$ is available in an \hrl\ $L_h$, then a multilingual model fine-tuned with $D^T_h$ is likely to transfer better to a related \lrl\ $L_i$ when $L_i$ and $L_h$ share several tokens in common. Thus, the objective  that \obpe\ seeks to optimize when creating a vocabulary is:
\begin{equation}
\begin{aligned}
\label{eq:obpe:goal}
    \cV =  & \argmin  \limits_{S: |S|=\vocabsize} \left[ (1-\alpha) \sum_{i=1}^n \lvert \text{encode}(D_{i},S) \rvert  \right. \\ &-\left. \alpha\sum\limits_{i\in \lset}\max\limits_{j\in \hset}\overlap(L_{i}, L_{j}, S) \right]
\end{aligned}
\end{equation}


where $0 \le \alpha \le 1$ determines importance of the two terms. The first term in the objective compactly represents the total corpus, as in BPE's (Eq~\eqref{eq:bpe:goal}).  The second term additionally biases towards vocabulary with greater overlap of each LRL to one \hrl\ where we expect task-specific labeled data to be present.  There are several ways in which we can measure the overlap between two languages with respect to a current vocabulary.  First, we encode each of $D_i$ and $D_j$ using the vocabulary $S$, which then yields a multiset of tokens in each corpus.  Inspired by the literature on fair allocation~\cite{barman2021universal}, we explore a continuously parameterized function that expresses overlap between two languages' encoding as a generalized mean function as follows: 
\begin{equation}
    \overlap(L_{i}, L_{h}, S) = \sum_{k \in S} \left(\frac{f_{ki}^{p} + f_{kh}^{p}}{2}\right)^\frac{1}{p},~~p \le 1
\label{Bilingual overlap defn}
\end{equation}
where $f_{ki}$ denotes  the frequency of token $k$ when $D_i$ is encoded with $S$. For different values of $p$, we get different tradeoffs between fairness to each language and overall goodness. When $p=-\infty$, generalized mean reduces to the minimum function, and we get the most egalitarian allocation.  However, this ignores the larger of the two frequencies.  When $p=1$, we get a simple average which is what the first term in Equation~\eqref{eq:obpe:goal} already covers. For $p=0,-1$, we get the geometric and harmonic means respectively. Due to smaller size of LRL monolingual data, the frequency of a token which is shared across languages is likely to be much higher in HRL monolingual data as compared to that in LRL monolingual data, Hence, setting $p$ to large negative values will increase the weight given to LRLs and thus increase overlap.  We will present an exploration of the effect of $p$ on zero-shot transfer in the experiment section.



\begin{table*}[h]
\begin{small}
    \centering
    \begin{tabular}{|l|l|l|r|r|} \hline
        Family & HRL & LRLs & \multicolumn{2}{|c|}{Number of HRL Docs} \\
         & & & \Balanced & \Skewed \\ \hline
                 West Germanic & English (en) & German (de), Dutch (nl), Western\ Frisian (fy) & 0.16M & 1.00M  \\ \hline
                 Romance & French (fr) & Spanish (es), Portuguese (pt), Italian (it)  & 0.16M & 0.50M \\ \hline
                  Indo-Aryan & Hindi (hi) & Marathi (mr), Punjabi (pa), Gujarati (gu) & 0.16M & 0.16M \\ \hline
    \end{tabular}
    \caption{Twelve Languages \emph{simulated} as HRLs and LRLs across with two different corpus distribution: \Balanced\ and \Skewed. Number of documents in languages simulated as LRLs is 20K.} 
    \label{tab:langs}
    \end{small}
\end{table*}
The greedy version of the above objective that controls the 
candidate vocabulary item to be 
inducted in each iteration of OBPE is thus: 
\begin{equation}
\label{eq:obpe:step}
\begin{split}
    \cV = \cV \cup \argmax_{k=[u,v]:u,v\in \cV} 
    (1-\alpha)\sum_{j}f_{kj} \\ + \alpha\sum_{i\in \lset}\max_{h \in \hset}\left(\frac{f_{ki}^{p} + f_{kh}^{p}}{2}\right)^\frac{1}{p}
\end{split}
\end{equation}
The data structure maintained by BPE to efficiently conduct such merges can be applied with little changes to the OBPE algorithm.  The only difference is that we need to separately maintain the frequency in each language in addition to overall frequency. Since the time and resources used to create the vocabulary is significantly smaller than the model pre-training time, this additional overhead to the  pre-training step is negligible.

\section{Experiments}
\label{sec:expts}

We evaluate by measuring the efficacy of zero-shot transfer from the \hrl\ on four different tasks: named entity recognition
(NER), part of speech tagging (POS), text classification(TC), and  Cross-lingual Natural Language Inference (XNLI).  
Through our experiments, we evaluate the following questions: 
\begin{enumerate}
    \item  Is OBPE more effective than BPE for zero-shot transfer? (\refsec{sec:effective-obpe})
    \item What is the effect of token overlap on overall accuracy? (\refsec{sec:analysis})
    \item How does increased LRL representation in the vocabulary impact accuracy? (\refsec{sec:samp})
    
    \end{enumerate}
We report additional ablation and analysis experiments in  \refsec{sec:ablation}.


\noindent
\subsection{Setup}
{\bf Pre-training Data and Languages}
As our pre-training dataset $\{D_i\}$,  we use 
the Wikipedia dumps of all the languages as used in mBERT. We pre-train with 12 languages grouped into three families of four related languages as shown in Table~\ref{tab:langs}. In each family, we simulate as HRL the most populous language, and call the remaining as LRLs. The number of documents for languages simulated as LRLs is set to 20K. For the HRLs, we consider two corpus distributions: 
\begin{itemize}
    \item \Balanced\ : all three HRLs get 160K documents each
    \item \Skewed\ : English gets one million, French half million, and Hindi 160K documents
\end{itemize}
We evaluate twelve-language models in each of these settings, and present results for separate four language models per family in Table \ref{tab:4_lang} in the Appendix.  For the Indo-Aryan languages set, the monolingual data of Punjabi and Gujarati is transliterated to Devanagari, the script of Hindi and Marathi. We use libindic’s indictrans library ~\cite{Bhat:2014:ISS:2824864.2824872} for
transliteration. Languages in the other two sets do not require transliteration as they have a common script. Thus, all four languages in each set are in the same script so their lexical overlap can be leveraged. 

\begin{table}[t]
    \centering
    \resizebox{\linewidth}{!}{
    \begin{tabular}{|l|l|r|r|r|r|}
    \hline
        \multirow{3}{*}{Dataset split} & Lang & \multicolumn{4}{|c|}{Number of sentences} \\ \hline
        ~ & ~ & NER & POS & TC & XNLI \\ \hline
        \multirow{3}{*}{Train:HRL} & hi & 5.0 & 53.0 & 25.0 & ~ \\ 
        & en & 10.5 & 18.0 & 10.0 & 393.0 \\ 
        & fr & 7.5 & 16.5 & 10.0 & 393.0 \\ \hline
        \multirow{3}{*}{Validation:HRL} & hi & 1.0 & 3.0 & 4.0 & \\
        ~ & en & 6.0 & 4.0 & 10.0 & 2.5 \\ 
        ~ & fr & 4.0 & 2.0 & 10.0 & 2.5 \\ \hline
        \multirow{3}{*}{Test data} & hi & 0.2 & 12.0 & 7.0 & ~ \\ 
        ~ & en & 6.0 & 4.6 & 10.0 & 5.0 \\ 
        ~ & fr & 4.0 & 4.1 & 10 & 5.0 \\ 
        ~ & mr & 0.8 & 9.5 & 6.5 & - \\ 
        ~ & pa & 0.2 & 13.4 & 7.9 & - \\ 
        ~ & gu & 0.3 & 14.0 & 8.0 & - \\
        ~ & de & 12.0 & 19.3 & 10.0 & 5.0 \\
        ~ & nl & 8.0 & 1.0 & - & - \\ 
        ~ & fy & 0.8 & - & - & - \\ 
        ~ & es & 5.0 & 3.1 & 10.0 & 5.0 \\ 
        ~ & pt & 4.0 & 2.5 & - & - \\
        ~ & it & 5.0 & 3.4 & - & - \\ \hline
    \end{tabular}
    }
    \caption{\label{tab:task:stats}Task-specific data sizes. Number of sentences in thousands.}
\end{table}
\noindent
{\bf Pre-Training Details}
To ensure that LRLs are not under-represented, we over-sample using exponentially smoothed weighting similar to multilingual BERT \cite{devlin-etal-2019-bert} with exponentiation factor 
0.7. We perform MLM pretraining on a BERT base model with 110M parameters from scratch. We generate a vocabulary of size of 30k.
We chose batch size as 2048, learning rate as 3e-5 and maximum sequence length as 128.  Pre-training of BERT
was done with duplication factor 5 for
for 64k iterations for HRLs.
For all LRLs, duplication factor was 20 and training was done for 24K iterations. MLM pre-training was done on Google
v3-8 Cloud TPUs where 10K iterations required 2.1 TPU hours.

\begin{table*}[!ht]
    \centering
    \begin{adjustbox}{max width=1.0\textwidth,center}
    \begin{tabular}{|l|l| l|l| l|l |l|l |l|}
    \hline
        \multirow{2}{*}{Method} & \multicolumn{4}{c|}{LRL Performance ($\uparrow$)} &\multicolumn{4}{c|}{HRL Performance ($\uparrow$)} \\ 
        & NER & TC & XNLI & POS & NER & TC & XNLI & POS \\
        \hline
        BPE \cite{sennrich-etal-2016-neural} & 64.48 & 65.52 & 52.07 & 84.64  & 83.26 
        & \textbf{82.07} 
        & 62.71 
        & \textbf{95.20} 
         \\ 
        BPE-dp \cite{provilkov-etal-2020-bpe} & 63.92 & 64.15 & 52.66 & 84.75  & 81.73 
        & 81.07 
        & 63.74 
        & 94.61
        \\
        
        CV \cite{chung-etal-2020-improving}  & 59.58  & 61.91  & 49.30  & 81.68 & 81.15 & 80.93& 64.51 & 94.47
        \\ 
        TokComp \cite{maronikolakis-etal-2021-wine-v}  & 63.79  & 65.77  & 53.94  & \textbf{85.49} & 82.43 & 80.93& 66.10 & 94.86
        \\ 
        {\method{} (This paper)}  & \textbf{65.72}& \textbf{68.02} & \textbf{54.03}  & 85.26 
        & \textbf{83.98}  & 81.91  & \textbf{66.27} & 95.09\\
    \hline
    \end{tabular}
    \end{adjustbox}
        \caption{Zero-shot performance of models in the Balanced-12 setting trained on 9 LRL and 3 HRL languages. Performance is measured on four tasks: NER (F1), Text Classification (Accuracy), POS (Accuracy), and XNLI (Accuracy). For all metrics, higher is better ($\uparrow$). Zero-shot transfer to LRL improves  without hurting \hrl\ accuracy. P-value of paired-t-test between BPE and OBPE LRL gains has values $0.01,0.04, 0.02,0.01$ for each of the 4 tasks establishing statistical significance. Detailed results for each language is pesented in Table~\ref{tab:varying_p}. \refsec{sec:effective-obpe} has further discussion.}
         \label{tab:overall} 
\end{table*}

\begin{table*}[htb]
    \centering
    \begin{adjustbox}{max width=1.1\textwidth,center}
    \begin{tabular}{|l|l| l|l| l|l |l|l |l|}
    \hline
        \multirow{2}{*}{Method} & \multicolumn{4}{c|}{LRL Performance ($\uparrow$)} &\multicolumn{4}{c|}{HRL Performance ($\uparrow$)} \\  
        & NER & TC & XNLI & POS & NER & TC & XNLI & POS \\
        \hline
        BPE \cite{sennrich-etal-2016-neural} & 52.91 & 51.68 & 48.57 & 74.79  & 81.78 
        & 80.04
        & 64.96 
        & 95.03
         \\ 
        CV \cite{chung-etal-2020-improving}  & 52.73 & 54.40 & 44.28  & 76.70
        & 79.84 & 77.74  & 57.18 & 94.60
        \\ 
        {\method{}\ (This paper)}  & {\bf 55.09} & {\bf 55.37} & {\bf 50.01}  & {\bf 75.05} & {\bf 82.94} & {\bf 80.31} & {\bf 65.57} & {\bf 95.09}\\
     \hline
    \end{tabular}
    \end{adjustbox}
        \caption{Zero-shot performance of models in the Skewed-12 setting of Table~\ref{tab:langs} on same four tasks as Table~\ref{tab:overall}. \method{} shows gains here too. Detailed numbers in Table~\ref{tab:12_lang:skew} of Supplementary.  \refsec{sec:effective-obpe} has further discussion.}
        \label{tab:overall:skew}
\end{table*}


\noindent
{\bf Task-specific Data} We evaluate on four down-stream tasks: (1) NER: data from WikiANN
\cite{pan-etal-2017-cross} and XTREME \cite{pmlr-v119-hu20b}, (2) XNLI: data from \cite{conneau-etal-2018-xnli},  (3) POS: data from XTREME \cite{pmlr-v119-hu20b} and TDIL\footnote{Technology Development for Indian Languages
(TDIL), https://www.tdil-dc.in},  and (4) Text Classification (TC): data from TDIL and XGLUE \cite{liang-etal-2020-xglue}. We downsampled the  
TDIL data for each
language to make them class-balanced. The POS
tagset for Indo-Aryan languages used was the BIS Tagset \cite{sardesai-etal-2012-bis}. Table~\ref{tab:task:stats} presents a summary.
The test set to compute LRL perplexity was formed by sampling 10K sentences from Samanantar corpus\cite{ramesh2021samanantar} for Indic languages and from Tatoeba corpus\footnote{Tatoeba
, https://tatoeba.org} for other languages. The perplexity reported for a language is the average of sentence perplexity over all the sentences sampled from that language's corpus.\\
\noindent
{\bf Task-specific fine-tuning details}
We perform task-specific fine-tuning of pre-trained BERT on the task-specific training data of \hrl\ and evaluate on all languages in the same family. Here we used
learning-rate 2e-5 and batch size 32, with training duration as 16 epochs for NER, 8 epochs for
POS and 3200 iterations for Text Classification and XNLI. The models were evaluated on a separate validation dataset of the HRL and the model with the minimum validation loss, maximum F1-score, accuracy and minimum validation loss was selected
for final evaluation for XNLI, NER, POS and Text Classification respectively. All fine-tuning experiments
were performed on Google Colaboratory. 
The results
reported for all the experiments are an average of 3
independent runs.

\subsection{Effectiveness of \method{}}
\label{sec:effective-obpe}
We evaluate the impact of \method{} on improving zero-shot transfer from HRLs to LRLs within the same family across four different tasks.  We compare with four 
existing methods that represent different methods of vocabulary creation and allocation of budget across languages:

\noindent
{\bf Methods compared}
\begin{enumerate}
    \item \textbf{BPE} \cite{sennrich-etal-2016-neural}, the existing default method of vocabulary generation.
    \item \textbf{Clustered vocabulary (CV)} \cite{chung-etal-2020-improving}  Since the paper uses a  SentencePiece unigram for vocabulary, we followed the same approach for this comparison.  We allocate each family equal number of vocabulary tokens which is \vocabsize/3.
    \item \textbf{BPE-dropout (BPE-dp)} \cite{provilkov-etal-2020-bpe} uses 
    the vocabulary generated by BPE but tokenizes the text using a dropout rate of 0.1.  This allows the training of tokens that are subsumed by larger tokens in the vocabulary.
    \item \textbf{Compatibility of Tokenizations (TokComp)} \cite{maronikolakis-etal-2021-wine-v}  uses a method to select meaningful vocabulary sizes in an automated manner for all language using compression rates. Since their best performances are found, when the compression rates are similar, we choose a size for each language corresponding to compression rate of 0.5. The tokenizer used in this method is WordPiece.
.
    \item \textbf{\method{}} (Ours) with default $\alpha=0.5, p = -\infty$. We also do ablation on these.
\end{enumerate}

In Table~\ref{tab:overall} we observe that across all four tasks, zero-shot LRL accuracy improves compared to BPE.  For example, the average accuracy on XNLI for the LRL languages improves from 55.6 to 58.1 just by changing the set of tokens in the vocabulary.  These gains are obtained without compromising HRL performance on the tasks. The Clustered Vocabulary (CV) approach is much worse than BPE. These experiments are on the Balanced-12 model.  In the supplementary section, we report the results on the Skewed-12 (Table~\ref{tab:overall:skew}) and Balanced-4 models (Table~\ref{tab:4_lang}) and show similar gains even with these models.
In this table, we averaged the gains over nine LRLs, and in the Supplementary Table~\ref{tab:varying_p} we show consistent gains for individual languages.

\begin{figure}[t]
  \centering
  \includegraphics[scale =0.32]{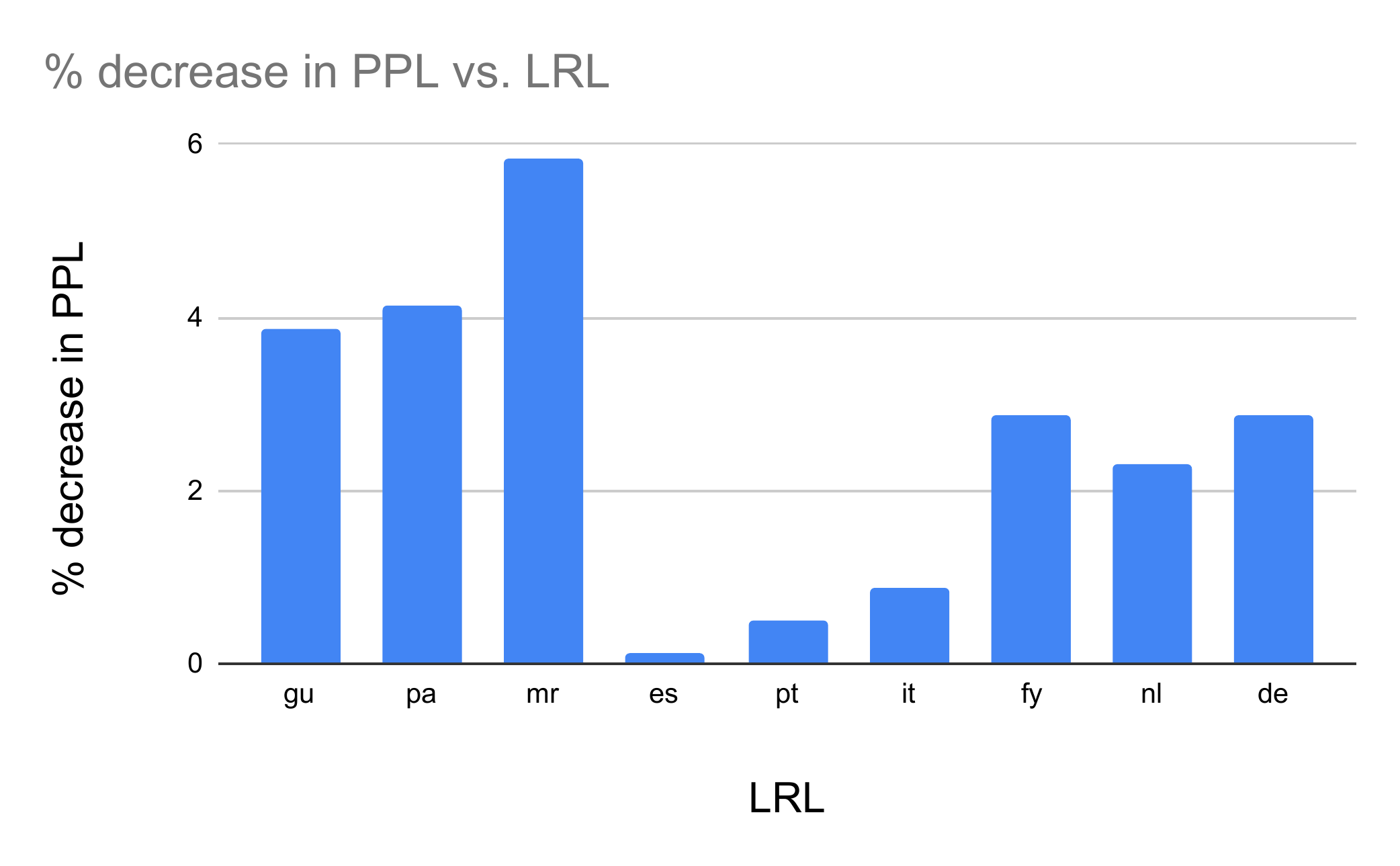}
  \caption{Percentage reduction in Pseudo perplexity~\cite{salazar-etal-2020-masked} for different LRLs as we go from BPE to \obpe\ vocabulary. (Section \ref{sec:effective-obpe})}
  \label{fig:ppl}
\end{figure}

In addition to improving zero-shot transfer from HRLs to LRLs on downstream tasks, \obpe\ also leads to better intrinsic representation of LRLs.  We validate that by measuring the pseudo-perplexity~\cite{salazar-etal-2020-masked} of a test set of LRL sentences.  We find that average perplexity of LRL sentences drops by 2.6\%
when we go from the BPE to OBPE vocabulary.  More details on this experiment appear in Figure~\ref{fig:ppl}. 

\begin{figure*}[ht]
\centering
\begin{tikzpicture}[scale=0.8]
\begin{groupplot}[group style={group size= 3 by 1,ylabels at=edge left},width=0.30\textwidth,height=0.25\textwidth]
\nextgroupplot[label style={font=\large},title=TC (Accuracy),
tick label style={font=\Large},
legend style={at={($(0,0)+(1cm,1cm)$)},legend columns=1,fill=none,draw=black,anchor=center,align=left,legend cell align=left,font=\small},
ylabel = {Performance},
legend to name=fred,
ymin=10,
ymax=100,
mark size=2pt]
\addplot [red,mark=square*] coordinates {(0, 84.50) (10, 77.16) (40, 73.43) (50, 79.81) (90, 81.96) (100, 84.56) };
\addplot [green,mark=*] coordinates {(0, 30.07) (10, 61.39) (40, 65.91) (50, 69.09) (90, 71.41) (100, 71.39) };

\addlegendentry{hi};    
\addlegendentry{mr};    
\coordinate (c1) at (rel axis cs:0,1);

\nextgroupplot[label style={font=\Large},title=NER (F1),
tick label style={font=\Large},
ymin=10,
ymax=100,
mark size=2pt]
\addplot [red,mark=square*] coordinates {(0, 85.14) (10, 82.04) (40, 82.45) (50, 83.01) (90, 84.84) (100, 86.34) };
\addplot [green,mark=*] coordinates {(0, 16.59) (10, 37.32) (40, 39.92) (50, 43.50) (90, 42.17) (100, 58.18) };
\nextgroupplot[label style={font=\Large},title=POS (Accuracy),
tick label style={font=\Large},
ymin=10,
ymax=100,
mark size=2pt]
\addplot [red,mark=square*]  coordinates {(0, 94.20) (10, 93.99) (40, 93.69) (50, 94.14) (90, 94.30) (100, 94.18)};
\addplot [green,mark=*] coordinates {(0, 74.07) (10, 80.79) (40, 78.78) (50, 78.75) (90, 79.16) (100, 81.92) };
\coordinate (c2) at (rel axis cs:1,1);

\end{groupplot}
\coordinate (c3) at ($(c2)$);
\node[right] at (c3 |- current bounding box.west)
{\pgfplotslegendfromname{fred}};

\end{tikzpicture}
\caption{Zero-shot performance vs Overlap 
of models trained on unicode shifted HRL data to simulate increasing overlap between HRL (SynthHindi) and LRL (mr). Performance is measured on three tasks: Text Classification (Accuracy), NER (F1) and POS (Accuracy).  On TC and NER observe the huge drop in LRL accuracy as we decrease overlap from 100 down to 0.  Further discussions in \refsec{sec:analysis}.}
\label{fig:unicode}

\end{figure*}
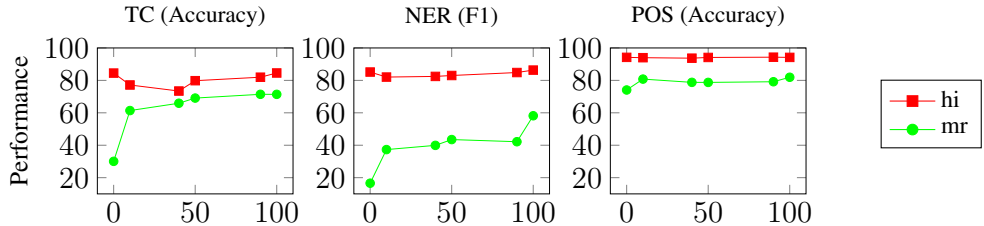

In order to investigate the reasons behind the OBPE gains, we first inspected the percentage of tokens in the vocabulary that belong to LRLs, HRLs, and in their overlap. We find that with OBPE both LRL tokens and overlapping tokens increase.  Either of these could have led to the observed gains.  We analyze the effect of each of these factors in the following two sections.

\subsection{Effect of Token Overlap}
\label{sec:analysis}

\begin{table}[!ht]
\begin{small}
    \centering
    \begin{tabular}{|l|c|c|}
    \hline
        ~  &\multicolumn{2}{c|}{en-es}  \\ \hline
        ~  & \multicolumn{1}{c|}{High (es: 1 GB)} &  \multicolumn{1}{c|}{Low: (es: 20K)}  \\ \hline
         NER & -1.4 & -11.7 \\ \hline
         XNLI & 0.7 & -1.3 \\ \hline
          ~  &\multicolumn{2}{c|}{hi-mr}  \\ \hline
        ~  & \multicolumn{1}{c|}{High (mr: 110K)} &  \multicolumn{1}{c|}{Low (mr: 20K)}  \\ \hline
         NER & -12.2  & -41.6 \\ \hline
         TC & -2.7 & -41.3 \\ \hline
         POS & -6.6 & -7.8 \\ \hline
    \end{tabular}
\caption{Drop in Accuracy of Zero-shot transfer when we synthetically reduce token overlap to zero.  Transfer is from English (en) as HRL to Spanish (es) and from Hindi (hi) as HRL to Marathi (mr) in two settings: (1) High where es, mr have sizes comparable to the HRL and (2) Low where their sizes are only 20K. Token overlap is important in the low-resource and related language setting (Section~\ref{sec:analysis})} 
\label{tab:overlapHL}
\end{small}
\end{table}

We present the impact of token overlap via two sets of experiments: first, a controlled setup where we synthetically vary the fraction of overlap and second where we measure correlation between overlap and gains of OBPE on the data as-is. 
For the controlled setup we follow \cite{K2020Cross-Lingual} for  synthetically controlling the amount of overlap between HRL and LRL. 
We trained a bilingual model between Hindi (HRL 160K) and Marathi (LRL 20K) --- two closely related languages in the Indo-Aryan family. 
To find the set of overlapping tokens between Hindi and Marathi, we first run \method{} on Hindi-Marathi language pair to generate a vocabulary and label all tokens present in both languages as 
\emph{overlapping tokens}. We then incrementally sample 10\%, 40\%, 50\%, 90\% of the tokens from this set. We shift the Unicode of the entire Hindi monolingual data except the set of sampled tokens so that there are no overlapping tokens between Hindi (hi) and Marathi (mr) monolingual data other than the sampled tokens. Let us call this Hindi data \textbf{SynthHindi}. We then run \method{} on SynthHindi-Marathi language pair to generate a vocabulary 
to pretrain the model. The task-specific Hindi data is also converted to SynthHindi during fine-tuning and testing of the model.

\reffig{fig:unicode} shows results with increasing overlap. We observe increasing gains in LRL accuracy as we go from no overlap to full overlap on all three tasks. NER accuracy increases from 17\% to 58\% for the LRL (mr) even while the HRL (hi) accuracy stays unchanged.  For TC we observe similar gains. For POS, even without token overlap, we get good cross-lingual transfer because POS tags are more driven by structural similarity, and Hindi and Marathi follow similar structure.



\begin{table*}[h]

\centering
\parbox{0.55\textwidth}{
\begin{small}
    \setlength\tabcolsep{3.0pt}
    \begin{tabular}{|l|l| l|l| l|l |l|l |l|}
    \hline
        \multirow{2}{*}{Method} & \multicolumn{4}{c|}{LRL Performance ($\uparrow$)} &\multicolumn{4}{c|}{HRL Performance ($\uparrow$)} \\ 
        & NER & TC & XNLI & POS & NER & TC & XNLI & POS \\
        \hline
        BPE  & 64.5 & 65.5 & 52.1 & 84.6  & 83.3 
        & \textbf{82.1} 
        & 62.7 
        & \textbf{95.2} 
         \\ 
        +overSample  & 64.4  & 67.6  & 52.1 & 84.6 & 82.4 & 82.0 & 62.0 & 95.2
        \\ \hline
        {\method{} }  & \textbf{65.7}& \textbf{68.0} & \textbf{54.0}  & \textbf{85.3} 
        & \textbf{84.0}  & 81.9  & \textbf{66.3} & 95.1\\
        +overSample   & 64.6  & 67.9  & 53.5  & 85.1 & 82.7 & 81.7 & 65.7 & 94.8
        \\ 
    \hline
    \end{tabular}
        \caption{\label{tab:sample}Zero-shot performance of models in the same setting as Table~\ref{tab:overall} but comparing default sampling with oversampling (exponentiation factor S=0.5). Note, even if BPE\_overSamp improves LRL somewhat, it causes HRL to drop. OBPE with default sampling is best for both LRLs and HRLs. Also OBPE\_overSampled is better than BPE\_overSampled (Section~\ref{sec:samp}). } 
\end{small}
}
\vspace{-0.6 cm}
\qquad
\begin{minipage}[c]{0.38\textwidth}%
\centering
\includegraphics[scale =0.3]{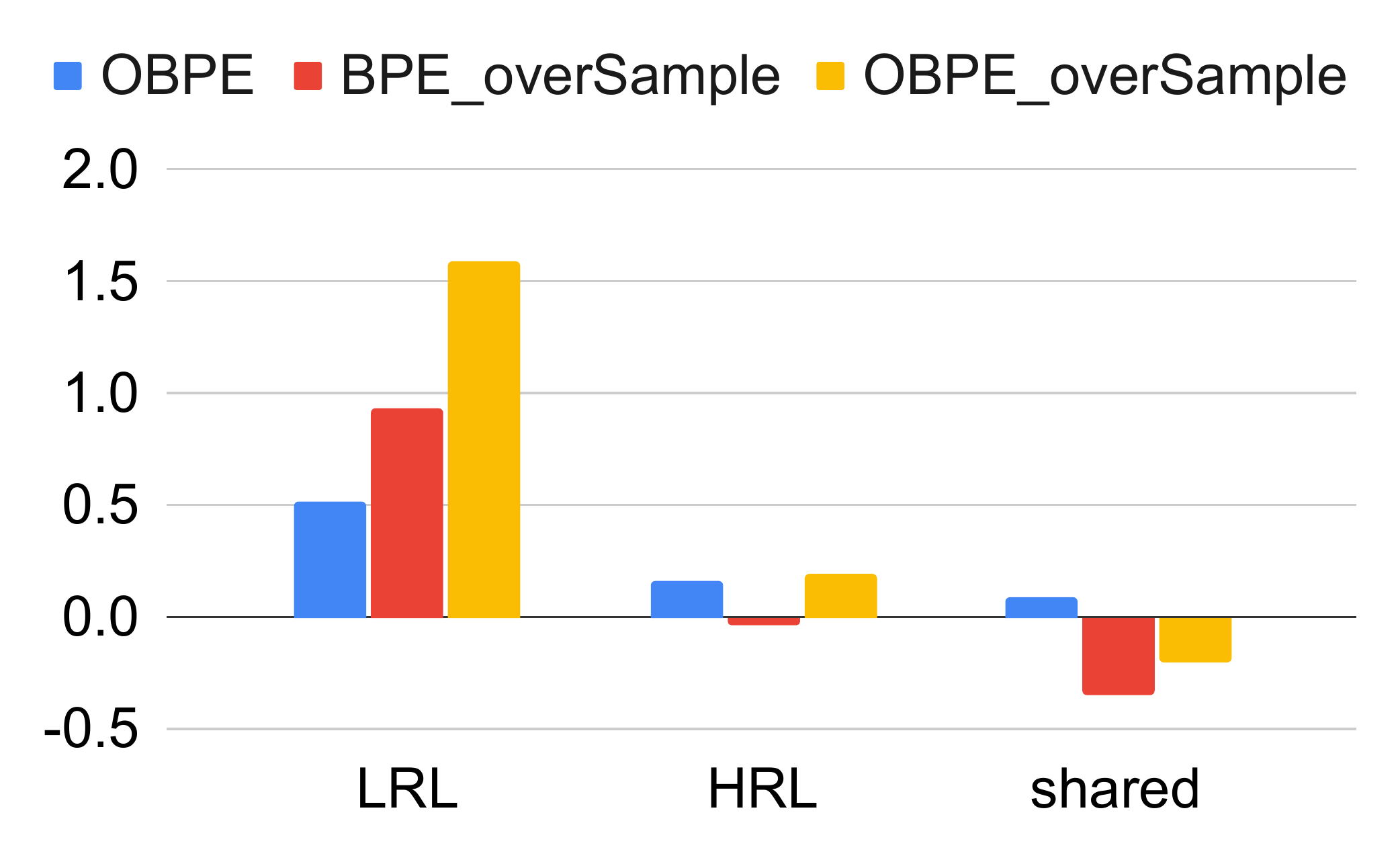}
\caption{Percentage rise over BPE in representation of LRL, HRL and Shared (percentage of tokens shared between HRL and LRL weighted by frequency) in vocabulary generated by OBPE and BPE\_overSample and OBPE\_overSample (Section~\ref{sec:samp}). }
\label{img:vocabstats}
\end{minipage}
\end{table*}

Our results contradict the conclusions of \cite{K2020Cross-Lingual} which claimed that token overlap is unimportant for cross-lingual transfer.  However, there are two key differences with our setting: (1) 
unlike \cite{K2020Cross-Lingual}, we explore low-resource settings, and (2) except for English-Spanish, the other language pairs they considered are not linguistically related. To explain the importance of both these factors,  in \reftbl{tab:overlapHL} we present accuracy of English-Spanish in a simulated low-resource setting where we sample 20K Spanish documents and 160K English documents. 
Also, we repeat our Hindi-Marathi experiments where Marathi is not low-resource.  We observe that
(1) Spanish as LRL benefits significantly on overlap with English.
(2) Marathi gains from token overlap with Hindi even in the high resource setting.

Thus, we conclude that as long as languages are related, token overlap is important and the benefit from overlap is higher in the low resource setting.

\paragraph{Overlap Vs Gain: Real data setup}
\label{sec:real}

\begin{table}[t]
\begin{small}
    \centering
    \begin{tabular}{|l|l|c|}
    \hline
        Lang family & Task & Pearson Correlation\\ \hline
        \multirow{2}{*}{Indo-Aryan} & NER & 0.835 \\ 
          & POS & 0.690 \\ \hline
         
        \multirow{2}{*}{West Germanic} & NER & 0.387 \\
          & POS & 0.348 \\ \hline
        \multirow{2}{*}{Romance} & NER & 0.946 \\ 
        & POS & 0.595 \\ \hline
    \end{tabular}
    \caption{\label{tab:corr}Correlation coefficient between performance gain and overlap gain within languages in a family for various tasks. 
    (Section~\ref{sec:real}).}
\end{small}
\end{table}
\vspace{-0.2cm}

We further substantiate our hypothesis that the shared tokens across languages favoured by \method{} enable transfer of supervision from HRL to LRL via statistics on real-data.  In Table~\ref{tab:corr} we show the Pearson product-moment correlation coefficient between overlap gain and performance gain within LRLs of the same family and task. We get a high positive correlation coefficient, with an average of 0.644. 

\subsection{Effect of Increased LRL representation}
\label{sec:samp}
We next investigate the impact of increased representation of LRL tokens in the vocabulary.  OBPE increases LRL representation by favoring overlapping tokens, but LRL tokens can also be increased by  just over-sampling LRL documents.  
We train another \Balanced 12 model but with further over-sampling LRLs with exponentiation factor of 0.5 instead of 0.7.  We observe in  Figure~\ref{img:vocabstats} that this increases LRL fraction but reduces HRL tokens in the vocabulary. Table~\ref{tab:sample} also shows the comparison of zero-shot transfer accuracy with over-sampled BPE against over-sampled OBPE. We find that OBPE even with default exponentiation factor achieves highest LRL gains, whereas aggressively over-sampled BPE hurts HRL accuracy. Within the same sampling setting, OBPE is better than corresponding BPE.

\subsection{Ablation study}
\label{sec:ablation}
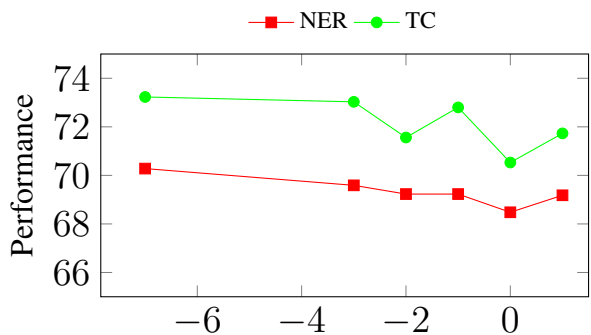
\begin{figure}[tbh]
\centering
\begin{tikzpicture}
\begin{groupplot}[group style={group size= 1 by 1,ylabels at=edge left},width=0.5\textwidth,height=0.3\textwidth]
\nextgroupplot[label style={font=\large},
tick label style={font=\Large},
tick label style={font=\Large},
legend style={at={($(1,-10)+(0cm,0cm)$),anchor=south},
legend pos=south west,
legend columns=2,fill=none,draw=none,anchor=south,align=left,legend cell align=left,font=\small},
ylabel = {Performance},
legend to name=fred,
xmin = =-inf,
xmax = 1.5,
ymin=65,
ymax=75,
mark size=2pt]
\addplot [red,mark=square*] coordinates {(-7,70.28)(-3,69.59)(-2,69.23)(-1,69.23)(0,68.48)(1,69.18)};
\addplot [green,mark=*] coordinates {(-7,73.23)(-3,73.03)(-2,71.56)(-1,72.80)(0,70.53)(1,71.73)};

\addlegendentry{NER};    
\addlegendentry{TC};    
       
\coordinate (c1) at (rel axis cs:0.5,0.0);
\coordinate (c3) at (rel axis cs:-1.0,1.0);

\end{groupplot}
\node[above] at (c1 |- current bounding box.north)
{\pgfplotslegendfromname{fred}};

\end{tikzpicture}
\caption{\label{fig:p}Zero-shot LRL performance of models in the same setting as Table~\ref{tab:overall} for different values of $p$ evaluated on NER and Text Classification. Best results at $p=-\infty$.(Section \ref{sec:ablation})}
\end{figure}

\vspace{-0.2cm}
We conducted experiments for different values of $p$ that controls the amount of overlap in the generalized mean function (\refeqn{eq:obpe:step}). Figure \ref{fig:p} and Table \ref{tab:varying_p} show the results for various $p$.  
Setting $p=1$  gives the original BPE algorithm. Setting $p=0,-1$ gives geometric and harmonic mean respectively, setting $p=-\infty$ gives minimum. We compare the task-specific results for different values of $p$ as shown in Table~\ref{tab:varying_p} and find that the gains we obtain are highest in the $p=-\infty$ (minimum) setting (Figure~\ref{fig:p}).

We also experiment with $\alpha = 0.7$, and find that for most languages the results were not better than our default  $\alpha = 0.5$. 


\section{Conclusion}

In this paper, we address the problem of  cross-lingual transfer from \hrl{}s to \lrl{}s by exploiting relatedness among them. We focus on lexical overlap during the vocabulary generation stage of multilingual pre-training. We propose Overlap BPE (\method{}), a simple yet effective modification to the BPE algorithm, which chooses a vocabulary that maximizes overlap across languages. \method{} encodes input corpora compactly while also balancing the
trade-off between cross-lingual subword sharing and language-specific vocabularies. We focus on three sets of closely related languages from diverse language families. Our experiments provide evidence that \method{} is effective in leveraging overlap across related languages to improve \lrl{} performance. In contrast to prior work, through controlled experiments on the amount of token overlap between two related \hrl{}-\lrl{} language pairs, we establish that token overlap is important when a LRL is paired with a related HRL.

\paragraph{Acknowledgements}
We thank Yash Khemchandani and Sarvesh Mehtani for participating in the early phases of this research.  We thank Dan Garrette and Srini Narayanan for comments on the draft.
We thank Technology Development for Indian Languages (TDIL) Programme initiated by the Ministry of Electronics
Information Technology, Govt. of India for providing us datasets used in this study. The experiments
reported in the paper were made possible by a Tensor Flow Research Cloud (TFRC) TPU grant. The
IIT Bombay authors thank Google Research India
for supporting this research.

\bibliography{custom,anthology} 
\bibliographystyle{acl_natbib.bst}


\appendix

\section{Appendix}
\label{sec:appendix}

\subsection{Examples of Token Overlap within Language Families}
\begin{figure}[h!]
  \centering
  \includegraphics[scale =0.30]{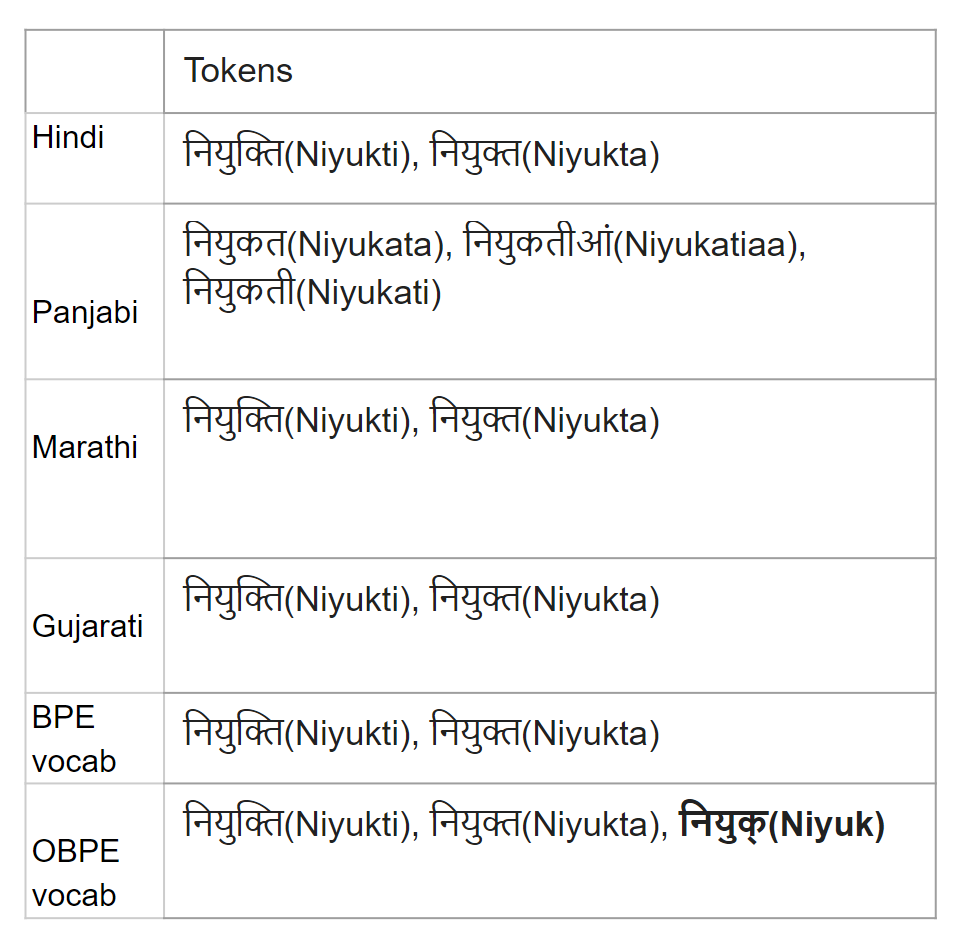}
  \caption{Similar meaning words with shared root forms across related Indo-Aryan languages. BPE vocabulary does not capture the tokens corresponding to Punjabi as it is a LRL and will thus tokenize Niyukata into multiple tokens which do not captures its meaning whereas Niyukata when tokenized by OBPE tokenizer will contain Niyuk which captures most of the meaning of the token Niyukata whose representation will be learnt when pretraining using Punjabi monolingual data }
  \label{fig:overlap-example}
\end{figure}

Table~\ref{tab:overlap-exampleAll} shows examples of overlapping tokens within three different language families, and 
Figure~\ref{fig:overlap-example} shows a real example of how OBPE chooses shared tokens.
\begin{table*}[]
\begin{small}
    \centering
    \begin{tabular}{|l||l|} \hline
        \multirow{2}{*}{Indo-Aryan} & Hindi:{\color{red}Vaapar}iyo, Marathi:{\color{red}Vaapar}tat , Punjabi:{\color{red}Vaapar}an, Gujarati:{\color{red}Vaapar}vana \\ 
          & Hindi:{\color{red}Jaa}te, Marathi:{\color{red}Jaa}oon , Punjabi:{\color{red}Jaa}na, Gujarati:{\color{red}Jaa}o \\ \hline\hline
        \multirow{2}{*}{West-Germanic} & English:C{\color{red}ategor}y, German:K{\color{red}ategor}ie, Dutch:C{\color{red}ategor}ie, Western Frisian:K{\color{red}ategor}y\\ 
         & English:{\color{red}Universit}y, German:{\color{red}Universit}aten, Dutch:{\color{red}Universit}eit, Western Frisian:{\color{red}Universit}eiten\\ \hline\hline
         \multirow{2}{*}{Romance} & French:{\color{red}Associa}tion, Spanish:{\color{red}Associa}cion, Portuguese:{\color{red}Associa}cao, Italian:{\color{red}Associa}zione\\ 
         & French:{\color{red}Certifi}e, Spanish:{\color{red}Certifi}car, Portuguese:{\color{red}Certifi}cado, Italian:{\color{red}Certifi}cato\\ \hline
        
    \end{tabular}
    \caption{\label{tab:overlap-exampleAll}Lexically overlapping tokens with similar meanings across four languages in each of three families. \method{}, our proposed method, exploits such meaning-preserving overlap among related languages to induce vocabulary for multilingual learning.}
    \end{small}
\end{table*}

\begin{table*}[t]
    \centering
    \tabcolsep 3pt
     \resizebox{\linewidth}{!}{
    \begin{tabular}{|l|l|l|l|l|l|l|l|l|l|l|l|l|l|l|}
    \hline
        Lang & hi & mr & pa & gu & en & de & nl & fy & fr & es & pt & it & LRL & HRL  \\ 
         & \hrl & & & & \hrl & & & & \hrl & & & & avg & avg\\
        \hline
        ~ & \multicolumn{14}{c|}{NER} \\ \hline
        BPE & 83.66 & 45.03 & 25.85 & 24.25 & 75.94 & 52.42 & 62.83 & 62.63 & 85.75 & 70.53 & 68.34 & 64.34 & 52.91 & 81.78 \\ 
        CV & 83.83 & 47.67 & 32.69 & 33.43 & 72.35 & 46.89 & 55.13 & 57.88 & 83.34 & 71.78 & 66.45 & 62.61 & 52.73 & 79.84\\
        \method{} & 85.92 & 47.55 & 26.05 & 32.79 & 77.15 & 52.72 & 62.87 & 65.55 & 85.76 & 73.35 & 70.25 & 64.69 & 55.09 & 82.94 \\ 
        \hline
        ~ & \multicolumn{14}{c|}{TC} \\ \hline
        BPE &  75.8 & 51.46 & 49.88 & 51.9 & 88.27 & 49.5 & ~ & ~ & 76.05 & 55.64 & ~ & ~ & 51.68 & 80.04\\
        CV & 76.46 & 54.37 & 55.49 & 56.33 & 81.94 & 51.5 & ~ & ~ & 74.81 & 54.31 & ~ & ~ & 54.40 & 77.74  \\
        \method{} &  76.58 & 55.38 & 53.98 & 54.06 & 88.3 & 57.85 & ~ & ~ & 76.06 & 55.59 & ~ & ~ & 55.37 & 80.31 \\ 
         \hline
     
        ~ & \multicolumn{14}{c|}{POS} \\ \hline
        BPE & 93.96 & 74.84 & 59.34 & 65.87 & 94.81 & 69.18 & 74.96 & ~ & 96.33 & 86.66 & 84.67 & 82.81 & 74.79 & 95.03  \\ 
        CV &  93.67 & 77.68 & 71.28 & 75.81 & 94.1 & 67.68 & 72.75 & ~ & 96.04 & 84.33 & 82.44 & 81.65 & 76.70 & 94.60 \\ 
        \method{} & 94.11 & 75.46 & 58.84 & 68.5 & 94.94 & 68.1 & 75.18 & ~ & 96.22 & 86.54 & 84.3 & 83.46 & 75.05 & 95.09  \\ \hline
      
        ~ & \multicolumn{14}{c|}{XNLI} \\ \hline
        BPE &~ & ~ & ~ & ~ & 67.05 & 45.51 & ~ & ~ & 62.87 & 51.62 & ~ & ~ & 48.57 & 64.96\\ 
        CV &~ & ~ & ~ & ~ & 54.87 & 39.87 & ~ & ~ & 59.48 & 48.68 & ~ & ~ & 44.28 & 57.18  \\ 
        \method{} &  ~ & ~ & ~ & ~ & 67.71 & 47.33 & ~ & ~ & 63.43 & 52.69 & ~ & ~ & 50.01 & 65.57 \\ \hline
    \end{tabular}}

    \caption{Zero-shot performance of models in the Skewed-12 setting  trained on 9 LRL and 3 HRL languages. Performance is measured on four tasks: NER (F1), Text Classification (Accuracy), POS (Accuracy), and XNLI (Accuracy). For all metrics, higher is better . Zero-shot transfer to LRL improves  without hurting \hrl\ accuracy. Averages results across HRLs and LRLs are presented in Table~\ref{tab:overall:skew}.  \method{} shows gains here too. \refsec{sec:effective-obpe} has further discussion.}

     \label{tab:12_lang:skew}
\end{table*}

\begin{table*}[t]
    \centering
    \tabcolsep 3pt
     \resizebox{\linewidth}{!}{
    \begin{tabular}{|l|l|l|l|l|l|l|l|l|l|l|l|l|l|l|l|l|}
    \hline
        Lang & hi & mr & pa & gu & en & de & nl & fy & fr & es & pt & it & LRL & HRL  \\ 
         & \hrl & & & & \hrl & & & & \hrl & & & & avg & avg\\
        \hline
        ~ & \multicolumn{14}{c|}{NER} \\ \hline
        BPE & 85.49 & 54.88 & 75.35 & 40.5 & 74.99 & 53.16 & 62.91 & 66.54 & 84.24 & 70.14 & 70.2 & 63.86 & 61.95 & 81.57 \\ 
        \method{}($\alpha$ = 0.5) & 86.59 & 59.23 & 76.15 & 41.84 & 74.74 & 56.95 & 63.19 & 67.92 & 83.73 & 69.99 & 69.76 & 64.91 & 63.33 & 81.69\\
      \method{}($\alpha$ = 0.7) & 85.99 & 59.54 & 75.59 & 41.37 & 75.36 & 54.6 & 63.43 & 66.86 & 83.95 & 71.77 & 69.27 & 66.29 & 63.19 & 81.77 \\ 
        \hline
        
        ~ & \multicolumn{14}{c|}{TC} \\ \hline
        bpe & 83.97 & 68.01 & 74.24 & 77.1 & 88.2 & 57.6 & ~ & ~ & 77.45 & 53.45 & ~ & ~ & 66.08 & 83.21\\
        \method{}($\alpha$ = 0.5) & 83 & 71.78 & 75.21 & 78.28 & 88.28 & 62.41 & ~ & ~ & 76.88 & 54.19 & ~ & ~ & 68.37 & 82.72  \\
        \method{}($\alpha$ = 0.7) & 83.56 & 69.3 & 74.84 & 77.09 & 87.93 & 57.9 & ~ & ~ & 77.11 & 57.84 & ~ & ~ & 67.39 & 82.87 \\ 
         \hline
     
        ~ & \multicolumn{14}{c|}{POS} \\ \hline
      bpe & 94.14 & 81.7 & 86.57 & 86.86 & 94.5 & 69.2 & 80.39 & ~ & 95.79 & 88.62 & 84.8 & 85.74 & 82.99 & 94.81   \\ 
      \method{}($\alpha$ = 0.5) & 94.18 & 82.79 & 86.63 & 86.5 & 94.6 & 70.53 & 79.49 & ~ & 95.94 & 88.79 & 86.62 & 86.41 & 83.47 & 94.91\\ 
      \method{}($\alpha$ = 0.7) & 94.1 & 81.56 & 87.04 & 86.55 & 94.38 & 70.67 & 79.99 & ~ & 96.17 & 89.8 & 87.77 & 86.19 & 83.70 & 94.88  \\ \hline
      
        ~ & \multicolumn{14}{c|}{XNLI} \\ \hline
        bpe &  ~ & ~ & ~ & ~ & 65.79 & 48.3 & ~ & ~ & 63.21 & 54.93 & ~ & ~ & 51.62 & 64.50\\ 
        \method{}($\alpha$ = 0.5) & ~ & ~ & ~ & ~ & 66.77 & 50.84 & ~ & ~ & 66.77 & 53.27 & ~ & ~ & 52.06 & 66.77  \\ 
        \method{}($\alpha$ = 0.7) &  ~ & ~ & ~ & ~ & 66.37 & 48.54 & ~ & ~ & 63.57 & 54.85 & ~ & ~ & 51.70 & 64.97 \\ \hline
    \end{tabular}}
    
    \caption{Zero-shot performance of three different models each trained on 3 LRLs and 1 HRL in the respective families \ref{tab:langs} in the \Balanced-4 setting  . Performance is measured on four tasks: NER (F1), Text Classification (Accuracy), POS (Accuracy), and XNLI (Accuracy). For all metrics, higher is better . Zero-shot transfer to LRL improves  without hurting \hrl\ accuracy.  \method{} shows gains here too. Languages in Romance family show some improvements in $\alpha$ = 0.7 setting as compared to $\alpha$ = 0.5. (Section {sec:ablation})}
     \label{tab:4_lang}
\end{table*}

\begin{table*}[tbh]
    \centering
    \begin{tabular}{|l|l|l|l|l|l|l|l|l|l|}
    \hline
        ~ & \multirow{2}{*}{\% overlap retained} &\multicolumn{4}{c|}{En-Es} &   \multicolumn{4}{c|}{Hi-Mr} \\ 
        ~ &  & \multicolumn{2}{c|}{High} &  \multicolumn{2}{c|}{Low} &  \multicolumn{2}{c|}{High} &  \multicolumn{2}{c|}{Low} \\ \hline
        ~ & ~ & es & en & es & hi & mr & hi & mr \\ \hline
        NER & 100 & 72.3 & 75.1 & 63.4 & 85.9 & 55.6 & 86.3 & 58.2 \\ 
        ~  & 0  & 70.9 & 67.7 & 51.7 & 82.7 & 43.4 & 85.1 & 16.6 \\ \hline
        TC & 100  & ~ & 88.2 & 63.7 & 84.4 & 75.1 & 84.6 & 71.4 \\
        ~  & 0  & ~ & 82.6 & 53.8 & 78.9 & 72.4 & 84.5 & 30.1 \\ \hline
        POS & 100  & ~ & 94.7 & 82.9 & 94.2 & 83.3 & 94.2 & 81.9 \\ 
        ~ & 0 &  ~ & 92.8 & 60.4 & 94.0 & 76.7 & 94.2 & 74.1 \\ \hline
        XNLI & 100  & 61.9 & 66.6 & 55.2 & ~ & ~ & ~ & ~ \\ 
        ~ & 0  & 62.6 & 61.5 & 53.9 & ~ & ~ & ~ & ~\\ \hline
    \end{tabular}
    \caption{Accuracy of Zero-shot transfer  from English (En) as HRL to Spanish (Es) and from Hindi(Hi) as HRL to Marathi(Mr) in two settings: (1) High where Es,Mr have sizes comparable to the HRL and (2) Low where their sizes are only 20K. As the percentage of overlapping tokens retained is decreased from 100\% to 0\%, the accuracy drops but the drop is higher in the low-resource setting. Task-specific accuracy numbers in first column(En-Es-High-es) have been taken from \cite{K2020Cross-Lingual}. Table \ref{tab:overlapHL} contains the reduction in accuracy on decreasing overlap from 100\% to 0 \% i.e. the difference between the rows corresponding to 100\% and 0\%}
    \label{tab:overlap:detailed}
\end{table*}

\begin{table*}[tbh]
\begin{small}
    \centering
    \begin{adjustbox}{max width=1.0\textwidth,center}
    \begin{tabular}{|l|l|l|l|l|l|l|l|l|l|l|l|l|l|}
    \hline
        Lang & hi & mr & pa & gu & en & de & nl & fy & fr & es & pt & it & avg  \\ \hline
        Method(p) & \multicolumn{13}{c|}{NER} \\ \hline
       OBPE(1)=BPE & 86.57 & 59.71 & 69.71 & 41.89 & 77.42 & 60.14 & 67.87 & 69.73 & 85.79 & 72.96 & 71.02 & 67.31 & 69.18 \\ 
       OBPE(0) & 87.29 & 61.86 & 67.21 & 41.46 & 76.08 & 59.50 & 67.30 & 66.86 & 86.02 & 69.80 & 70.89 & 67.54 & 68.48 \\ 
        OBPE(-1) & 86.67 & 64.19 & 72.38 & 39.93 & 77.17 & 58.25 & 67.09 & 69.86 & 85.83 & 72.99 & 70.43 & 66.02 & 69.23 \\
        OBPE(-2) & 86.17 & 60.91 & 67.30 & 44.43 & 76.47 & 59.66 & 67.13 & 70.03 & 85.25 & 75.02 & 71.82 & 66.53 & 69.23 \\
         OBPE(-3) & 87.14 & 62.68 & 72.25 & 44.73 & 77.24 & 61.41 & 67.38 & 69.87 & 86.15 & 69.82 & 71.06 & 65.37 & 69.59 \\ 
         OBPE(-$\infty$) & 87.09 & 62.96 & 72.17 & 44.25 & 77.93 & 60.44 & 68.65 & 70.23 & 86.92 & 74.14 & 72.55 & 66.05 & 70.28 \\
         BPE-dp & 85.54 & 62.64 & 71.46 & 39.75 & 75.51 & 59.29 & 67.76 & 70.42 & 84.15 & 67.43 & 68.82 & 67.74 & 68.38 \\ 
         TokComp & 86.43 & 61.12 & 72.82 & 45.88 & 76.57 & 55.25 & 65.28 & 67.85 & 84.22 & 71.04 & 68.87 & 66.00 & 68.44 \\
         CV & 84.27 & 55.66 & 43.37 & 50.19 & 74.99 & 53.51 & 65.36 & 65.39 & 84.20 & 73.05 & 66.16 & 63.49 & 64.97 \\
        Bsamp & 84.68 & 59.73 & 67.31 & 40.62 & 76.76 & 61.34 & 67.29 & 71.80 & 85.89 & 73.71 & 71.20 & 66.69 & 68.92 \\ 
        Osamp & 84.71 & 63.22 & 67.82 & 42.03 & 77.83 & 62.35 & 68.08 & 71.59 & 85.50 & 69.16 & 70.06 & 66.70 & 69.09 \\
        \hline
        ~ & \multicolumn{13}{c|}{TC} \\ \hline
         OBPE(1)=BPE & 80.35 & 61.45 & 69.00 & 72.32 & 88.63 & 62.27 & ~ & ~ & 77.23 & 62.58 & ~ & ~ & 71.73 \\ 
        OBPE(0) & 80.11 & 64.07 & 68.26 & 70.48 & 87.61 & 54.96 & ~ & ~ & 76.53 & 62.23 & ~ & ~ & 70.53 \\ 
        OBPE(-1) & 80.00 & 64.37 & 69.10 & 72.10 & 87.89 & 66.25 & ~ & ~ & 77.33 & 65.36 & ~ & ~ & 72.80 \\ 
        OBPE(-2) & 79.21 & 64.83 & 68.58 & 70.41 & 88.17 & 65.76 & ~ & ~ & 76.78 & 58.71 & ~ & ~ & 71.56\\
        OBPE(-3) & 81.00 & 62.79 & 68.17 & 73.20 & 89.38 & 68.34 & ~ & ~ & 77.50 & 63.84 & ~ & ~ & 73.03 \\
        OBPE($-\infty$) & 80.68 & 68.90 & 70.03 & 72.14 & 87.92 & 66.05 & ~ & ~ & 77.14 & 63.00 & ~ & ~ & 73.23 \\ 
        BPE-dp & 79.68 & 63.45 & 69.43 & 70.36 & 87.39 & 59.75 & ~ & ~ & 76.15 & 57.76 & ~ & ~ & 70.50 \\
        TokComp & 82.06 & 67.17 & 70.42 & 72.48 & 88.02 & 58.47 & ~ & ~ & 77.22 & 60.29 & ~ & ~ & 72.02 \\ 
        CV & 79.90 & 61.33 & 65.68 & 68.96 & 87.98 & 55.79 & ~ & ~ & 74.92 & 57.79 & ~ & ~ & 69.04 \\ 
        Bsamp & 81.00 & 65.29 & 70.97 & 72.30 & 88.05 & 66.11 & ~ & ~ & 76.92 & 63.51 & ~ & ~ & 73.02 \\ 
        Osamp & 80.11 & 66.08 & 70.11 & 72.38 & 88.39 & 66.25 & ~ & ~ & 76.57 & 64.58 & ~ & ~ & 73.06 \\
        \hline
        ~ & \multicolumn{13}{c|}{POS} \\ \hline
           OBPE(1)=BPE & 94.22 & 79.60 & 86.83 & 86.21 & 94.91 & 77.70 & 82.00 & ~ & 96.47 & 89.74 & 87.79 & 87.27 & 87.52 \\ 
        OBPE(0) & 94.13 & 76.26 & 86.53 & 85.03 & 94.85 & 76.09 & 82.48 & ~ & 96.31 & 88.78 & 87.01 & 86.62 & 86.74 \\ 
        OBPE(-1) & 94.20 & 79.13 & 86.23 & 85.14 & 94.87 & 78.22 & 82.56 & ~ & 96.32 & 89.62 & 87.25 & 87.27 & 87.34 \\ 
        OBPE(-2) & 93.98 & 81.07 & 86.54 & 85.86 & 94.68 & 76.80 & 82.08 & ~ & 96.23 & 89.14 & 86.31 & 86.45 & 87.19  \\
        
        OBPE(-3) & 94.31 & 79.55 & 86.67 & 86.65 & 95.03 & 76.34 & 83.63 & ~ & 96.30 & 89.97 & 87.76 & 88.00 & 87.66 \\
          OBPE($-\infty$) & 94.18 & 81.55 & 87.01 & 86.76 & 94.98 & 79.28 & 82.38 & ~ & 96.40 & 90.04 & 88.01 & 88.21 & 87.94 \\
          BPE-dp & 93.26 & 80.03 & 86.31 & 85.23 & 94.49 & 77.90 & 83.07 & ~ & 96.10 & 90.01 & 87.84 & 87.63 & 87.44 \\
          TokComp & 93.99 & 80.38 & 86.75 & 86.79 & 94.79 & 79.50 & 84.91 & ~ & 95.80 & 89.87 & 87.05 & 88.70 & 88.05 \\
          CV & 93.12 & 74.82 & 84.62 & 81.56 & 94.28 & 74.74 & 79.41 & ~ & 96.01 & 87.51 & 85.52 & 85.27 & 85.17 \\
        Bsamp & 94.3 & 77.81 & 86.35 & 85.68 & 94.93 & 77.2 & 82.94 & ~ & 96.34 & 89.93 & 88.05 & 88.54 & 87.46 \\ 
        Osamp & 94.01 & 81.22 & 86.66 & 86.36 & 94.35 & 76.99 & 82.93 & ~ & 96.04 & 89.75 & 88.24 & 88.43 & 87.73 \\

    \hline
        ~ & \multicolumn{13}{c|}{XNLI} \\ \hline
        OBPE(1)=BPE & ~ & ~ & ~ & ~ & 64.35 & 50.36 & ~ & ~ & 61.06 & 53.77 & ~ & ~ & 57.39 \\ 
        OBPE(0) & ~ & ~ & ~ & ~ & 64.33 & 49.06 & ~ & ~ & 59.96 & 54.71 & ~ & ~ & 57.02 \\ 
        OBPE(-1) & ~ & ~ & ~ & ~ & 64.35 & 48.62 & ~ & ~ & 61.40 & 53.51 & ~ & ~ & 56.97 \\ 
        OBPE(-2) & ~ & ~ & ~ & ~ & 65.05 & 50.36 & ~ & ~ & 64.45 & 55.31 & ~ & ~ & 58.79 \\ 
        OBPE(-3) & ~ & ~ & ~ & ~ & 67.86 & 50.64 & ~ & ~ & 64.85 & 57.11 & ~ & ~ & 60.11 \\         
        OBPE($-\infty$) & ~ & ~ & ~ & ~ & 67.41 & 50.76 & ~ & ~ & 65.13 & 57.29 & ~ & ~ & 60.15 \\  
        BPE-dp & ~ & ~ & ~ & ~ & 64.31 & 50.16 & ~ & ~ & 63.17 & 55.17 & ~ & ~ & 58.20 \\
        TokComp & ~ & ~ & ~ & ~ & 67.98 & 53.05 & ~ & ~ & 64.21 & 54.83 & ~ & ~ & 60.02 \\
        CV & ~ & ~ & ~ & ~ & 65.19 & 47.43 & ~ & ~ & 63.83 & 51.16 & ~ & ~ & 56.90 \\ 
        Bsamp & ~ & ~ & ~ & ~ & 63.41 & 51.02 & ~ & ~ & 60.58 & 53.09 & ~ & ~ & 57.03 \\ 
        Osamp & ~ & ~ & ~ & ~ & 67.13 & 50.38 & ~ & ~ & 64.25 & 56.64 & ~ & ~ & 59.60 \\ \hline
        
\hline
    \end{tabular}
    \end{adjustbox}
    \caption{Zero-shot performance of models in the Balanced-12 setting trained on 9 LRL and 3 HRL languages. Performance is measured on four tasks: NER (F1), Text Classification (Accuracy), POS (Accuracy), and XNLI (Accuracy). For all metrics, higher is better . Zero-shot transfer to LRL improves  without hurting \hrl\ accuracy. Averages results across HRLs and LRLs are presented in Table~\ref{tab:overall}. \refsec{sec:effective-obpe} has further discussion.Table \ref{tab:overall} contains the values corresponding to rows BPE, BPE-dp, CV, TokComp, OBPE($-\infty$) averaged over LRLs and HRLs, Table \ref{tab:sample} contains the values corresponding to rows Bsamp, Osamp averaged over LRLs and HRLs, , Figure \ref{fig:p} plots the rows correponding to varying p values. (Section \ref{sec:ablation})}
    \label{tab:varying_p}
    \end{small}
\end{table*}

\subsection{Limitations}
\begin{itemize}
    \item Our approach is expected to improve cross-lingual transfer from HRL to LRL only when the HRL and LRL are related linguistically since it relies on the presence of lexically overlapping tokens
    \item It requires the transliteration of LRL data to the script of its related HRL if LRL does not have the same script. 
\end{itemize}

\subsection{Potential risks}
    Language models may amplify bias in data and also introduce new ones. Multilingual models explored in the paper are not immune to such issues. Detecting such biases and mitigating them is a topic of ongoing research. We are hopeful that our focus on better representation of LRLs in the vocabulary is a step towards more inclusive models.

\subsection{Replicability}

    BERT configuration parameters used in our experiments are as follows:\\
    
 { "attention\_probs\_dropout\_prob": 0.1,
  "hidden\_act": "gelu",
  "hidden\_dropout\_prob": 0.1,
  "hidden\_size": 768,
  "initializer\_range": 0.02,
  "intermediate\_size": 3072,
  "max\_position\_embeddings": 512,
  "num\_attention\_heads": 12,
  "num\_hidden\_layers": 12,
  "type\_vocab\_size": 2,
  "vocab\_size": 30000
}
All the task-specific fine-tuning experiments are done using GPUs on Google Colaboratory where each fine-tuning experiment requires 2 GPU hours.

\subsection{License}

   Tatoeba data, GLUE data, Wikipedia dumps use the Creative Commons licenses.
   TDIL data used for Indic languages uses Research license type and Xtreme dataset uses Apache License 2.0. To the best of our knowledge, the use of scientific artifacts in this work is consistent with their intended use.
    
\subsection{Data bias}
We have used standard Wikipedia corpus, and there have been some studies on bias in such corpus.\cite{10.1145/3041021.3053375}

\end{document}